\documentclass[runningheads]{llncs}
\usepackage{graphicx}

\usepackage{tikz}
\usepackage{comment}
\usepackage{amsmath,amssymb} 
\usepackage{color}

\usepackage{algorithm}
\usepackage{algpseudocode}

\usepackage[accsupp]{axessibility}  

\usepackage{xspace}
\newcommand*{\eg}{e.g.\@\xspace}
\newcommand*{\ie}{i.e.\@\xspace}
\newcommand*{\etal}{et al.\@\xspace}

\newcommand*{\wrt}{w.r.t. \@\xspace}
\newcommand*{\cf}{cf. \@\xspace}

\usepackage{multirow}
\usepackage{array}
\newcolumntype{C}[1]{>{\centering\arraybackslash}p{#1}}
\usepackage{subcaption}

\usepackage[pagebackref,breaklinks,colorlinks]{hyperref}

\begin{document}
\pagestyle{headings}
\mainmatter
\def\ECCVSubNumber{4940}  

\title{Few `\textit{Zero Level Set}'-Shot Learning of Shape Signed Distance Functions in Feature Space
} 


\titlerunning{Few `\textit{Zero Level Set}'-Shot Learning of SDFs in Feature Space}
%
\author{\large Amine Ouasfi \quad
Adnane Boukhayma}
\authorrunning{Amine Ouasfi, Adnane Boukhayma}
%
\institute{Inria, Univ. Rennes, CNRS, IRISA, M2S, France}
\maketitle

\begin{abstract}
We explore a new idea for learning based shape reconstruction from a point cloud, based on the recently popularized implicit neural shape representations. We cast the problem as a few-shot learning of implicit neural signed distance functions in feature space, that we approach using gradient based meta-learning.
We use a convolutional encoder to build a feature space given the input point cloud. An implicit decoder learns to predict signed distance values given points represented in this feature space. Setting the input point cloud, \ie samples from the target shape function's zero level set, as the support (\ie context) in few-shot learning terms, we train the decoder such that it can adapt its weights to the underlying shape of this context with a few  (5) tuning steps. We thus combine two types of implicit neural network conditioning mechanisms simultaneously for the first time, namely feature encoding and meta-learning. Our numerical and qualitative evaluation shows that in the context of implicit reconstruction from a sparse point cloud, our proposed strategy, \ie meta-learning in feature space, outperforms existing alternatives, namely standard supervised learning in feature space, and meta-learning in euclidean space, while still providing fast inference.
\end{abstract}

\section{Introduction}
\label{sec:intro}

One of the driving motives behind the ongoing research in 3D computer vision is enabling machines to reason about and understand 3D given limited observations in the same way we humans can evidently do. This ability is in turn crucial for most downstream 3D based computer vision and machine learning tasks. A popular instance of this ability is manifested in the problem of full 3D shape reconstruction from a sparse incomplete point cloud. The prominence of this problem is additionally due to the ubiquity of such partial inputs, either as acquired through the increasingly accessible 3D scanning technologies, or being an intermediate output of numerous classical computer vision algorithms such as Structure from Motion or Multi-View Stereo. Classical solutions to this problem such as Poisson surface reconstruction \cite{kazhdan2013screened} still offer competitive reconstruction performances from dense point sets. However, as the inputs get sparser and less complete, learning based approaches become naturally more suitable to the task by virtue of their capacity to reason about shapes more globally and inpaint missing information based on previously seen examples. 

A class of the these learning based approaches that emerged recently proposes to represent shapes in the form of an implicit function whose zero level set coincides with the surface, parameterised by a neural network. Compared to their traditional alternatives, these representations offer many advantages, most notably enabling modelling shapes with variable topology unlike point clouds and meshes, and operating virtually at infinite spatial resolution unlike voxel grids. In practice, these shape functions are typically multi layer perceptrons mapping the domain to the co-domain, \ie 3D euclidean space to occupancies or signed distances. The zero level set of the inferred field can be rendered differentiably through \eg variants of ray marching \cite{hart1996sphere} and tessellated into explicit meshes with \eg Marching Cubes \cite{lorensen1987marching}. Coupling these implicit neural functions with a conditioning mechanism allows generalization across multiple shapes. For instance, combining their inputs with features generated from an additional encoder network yields single forward pass inference models that can learn to reconstruct from various input modalities. In particular,  recent models \cite{chibane2020implicit,peng2020convolutional} obtaining state-of-the-art performances on reconstruction from point cloud benchmarks \cite{shapenet}  
use a convolutional encoder that builds a feature embedding for euclidean points given the input point cloud. The implicit neural shape function learns to map these points from that feature space to their occupancy or signed distance values. These models are trained using dense points sampled near the surface with corresponding ground-truth singed distance or occupancy values. Our aim here is to improve the performance of such models with negligible additional test-time computational cost.    

As obtaining larger training data corpora remains prohibitive in 3D, most recent advances in this avenue focus on revamping the models, \eg their architectures \cite{sitzmann2020implicit}, input representations \cite{tancik2020fourier,chibane2020implicit,peng2020convolutional}, training objectives \cite{gropp2020implicit,lipman2021phase}, and training procedures \cite{duan2020curriculum}, while remaining within the standard supervised learning paradigm. Conversely, we propose here to cast the problem of surface reconstruction from a point cloud, with an encoder endowed implicit neural function, as a few-shot learning problem.

Beyond merely using the input point cloud in a single encoding forward pass for inference, 
we observe that we can additionally further fine-tune the conditioned shape function using the point cloud elements as training samples \cite{gropp2020implicit}, as they naturally belong to the surface and hence can be used to further overfit the shape signed distance function with their zero target values. To ensure this fine-tuning improves the initial result and that it is initialized from optimal shape function weights, we formalize it in a more principled learning strategy that is few-shot learning \cite{vinyals2016matching,snell2017prototypical,finn2017model}. Each shape is represented by a support set: the points of the input point cloud, and a query set: the dense pre-sampled training points. For a given shape, the objective is to optimize predictions on the query set, \ie adapt the shape function to the current shape, using the support set. We implement this strategy using gradient based meta-learning, namely the MAML algorithm \cite{finn2017model}. At every training step on a given shape, the adaptation consists in back-propagating the loss on the sparse support at the surface for a few iterations (5 steps). The main shape function's parameters are then updated by back-propagating the loss of the adapted shape function on the dense query set. Notice that by representing points in feature space during this process, we combine two types of implicit shape function conditioning through both the encoder and meta-learning for the first time.  

Using standard test beds we show that our approach 
outperforms comparable baselines in various 3D shape reconstruction metrics, and we provide qualitative results that support this as well. Through our experiments, we show that using few-shot leaning in feature space improves on both standard supervised learning in feature space (IF-Nets \cite{chibane2020implicit}) and few-shot learning in euclidean space (MetaSDF \cite{sitzmann2020metasdf}), both in single and multi-class shape setups for shape reconstruction from a sparse point cloud (less than 3k input points). The performance gap \wrt our standard supervised learning baseline increases even further with coarser inputs. We note that we follow the same experimental data setup as in our baselines IF-Nets and MetaSDF. We also point that while we use IF-Nets as our backbone model in this work, this idea could be extended to any convolutional encoder equipped implicit neural shape network.\\

\section{Related Work}

We review in this section work that we deemed most relevant to the context of our contribution.\\

\vspace{-10pt}
\noindent\textbf{Traditional Shape Representations}\quad
Perhaps an intuitive way to categorize 3D shape representations  within deep learning frameworks is into intrinsic and extrinsic representations. Intrinsic representations are efficient in that they are discretizations of the shape itself. however when represented explicitly, as in meshes \cite{wang2018pixel2mesh,kato2018neural} or point clouds \cite{fan2017point}, they are inherently limited to a fixed topology, which is unpractical for generating varying shape objects and classes.
Other forms of intrinsic representations include combining 2D patches \cite{groueix2018papier,williams2019deep,deprelle2019learning}, 3D shape primitives such as cuboids \cite{abstractionTulsiani17,zou20173d}, planes \cite{liu2018planenet} and Gaussians \cite{genova2019learning}. However patches induce discontinuities, and the approximation quality of primitive shapes remains limited by their simplicity. Extrinsic representations on the other hand model the 3D space containing the shape of interest. The most adopted one to date has been voxel grids \cite{wu20153d,wu2016learning}, being a natural extension of 2D pixels to 3D. Nonetheless, the cubic memory cost in voxel grid resolution limits the ability to represent details. Sparse voxel representations such as octrees \cite{riegler2017octnet,tatarchenko2017octree,wang2017cnn} can help alleviate these memory efficiency issues albeit with complex implementations.\\     

\vspace{-10pt}
\noindent\textbf{Implicit Neural Shape Representations}\quad
Recent years have seen a surge in extrinsic implicit neural shape representations for modelling 3D objects and scenes. Thanks to their ability to continuously represent detailed shapes with arbitrary topologies in a memory-efficient way, these representations remedy many of the shortcomings of the aforementioned traditional alternatives, and are currently drawing increasing attention both in 3D shape and appearance modelling (\eg \cite{mildenhall2020nerf,kellnhofer2021neural,NEURIPS2020_1a77befc}).
Implicit neural shape models are typically parameterized with MLPs that map 3D space to occupancy \cite{mescheder2019occupancy}, signed \cite{park2019deepsdf} or unsigned distances \cite{chibane2020neural} relative to the shape. 
Different forms of training supervision have been proposed, the most common one being 3D points pre-sampled around the surface, with weaker forms of supervision such as 2D segmentation masks through a 2D based SDF lower bound \cite{lin2020sdf}, or color and depth images \cite{niemeyer2020differentiable,kellnhofer2021neural,NEURIPS2020_1a77befc} through differentiable rendering \cite{jiang2020sdfdiff,NEURIPS2019_bdf3fd65,jiang2020sdfdiff}. Recent contributions in this area include learning octree scafolded implicit mooving least squares  \cite{liu2021deep}, representing shapes as an implicit template and an implicit warp \cite{zheng2021deep}, and implicit/explicit hybrid representations \cite{chen2020bsp,deng2020cvxnet,yavartanoo20213dias} based on differentiable space partitioning.\\ 

\vspace{-10pt}
\noindent\textbf{Conditioning Implicit Neural Shape Models}\quad
Implicit shape models require conditioning mechanisms to represent more than a single shape. The mechanisms explored so far include concatenation, batch normalization, hypernetworks \cite{sitzmann2020implicit,NEURIPS2019_b5dc4e5d,sitzmann2021light}
and meta-learning \cite{sitzmann2020metasdf}. 
Concatenation like conditioning was first introduced through a single latent code \cite{mescheder2019occupancy,chen2019learning,park2019deepsdf}, and subsequently improved through the use of local features \cite{genova2020local,tretschk2020patchnets,takikawa2021neural,peng2020convolutional,chibane2020implicit,jiang2020local}. 

Current methods that meta-learn implicit 3D neural representations use gradient based meta-learning (\eg MAML \cite{finn2017model}, Reptile \cite{nichol2018first}) to learn a meta-radiance field that can be adapted from images \cite{tancik2021learned,gao2020portrait}, or a meta-SDF that can be adapted from both zero level set and random domain samples \cite{sitzmann2020metasdf}. In contrast to these methods,  we propose here to combine encoder-based local feature concatenation conditioning and meta-learning conditioning in the same model, performing implicit reconstruction from a sparse point cloud. 
We note also that in the work by Sitzmann \etal, the meta-learning conditioning requires many surface samples (10k new points sampled at each of the 5 MAML iterations = 50k pts). Differently, we extend this idea to a true few-shot reconstruction setup (300 or 3k fixed input points) and multi shape class for the first time, and show that it can only scale thusly in feature space.


\begin{figure*}[t!]
\centering
\includegraphics[width=0.9\linewidth]{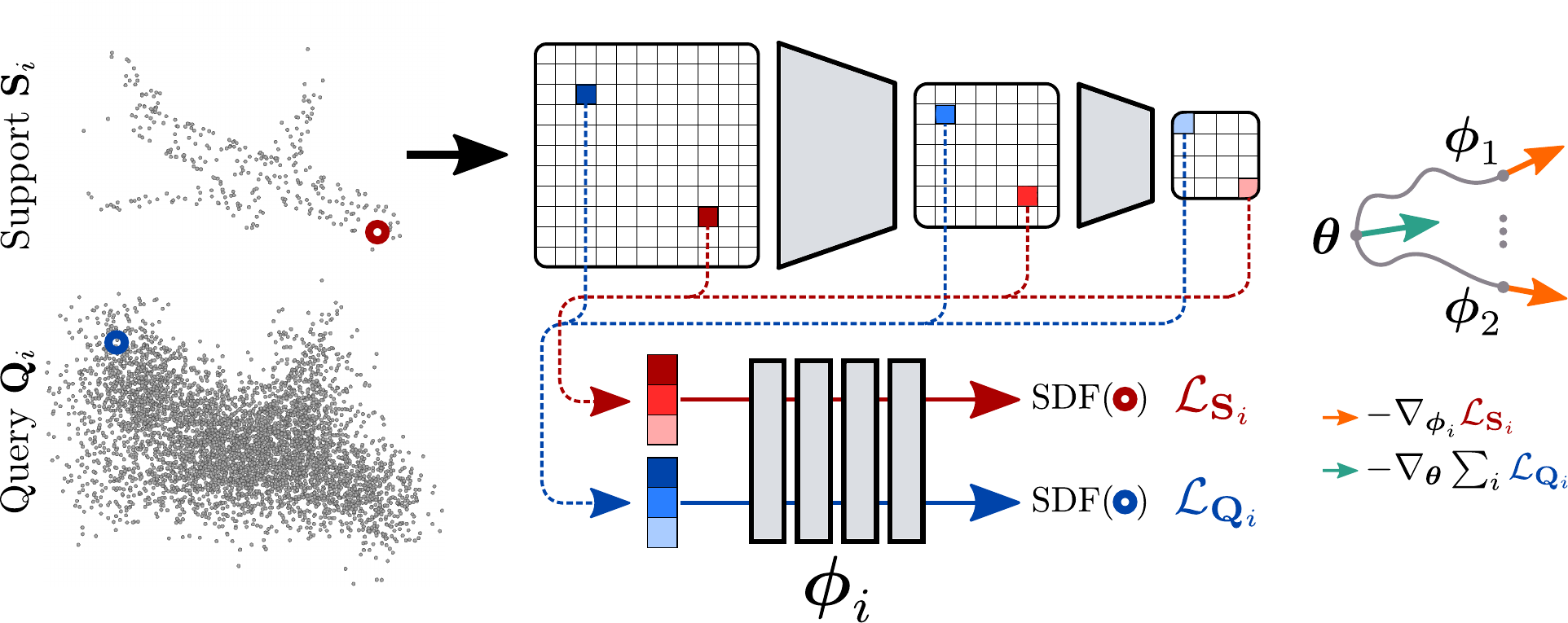}
\vspace{-12pt}
  \caption{\small Overview of our method. \quad Our input is a sparse point cloud (Support $\mathbf{S}_i$) and our output is an implicit neural SDF $f$. $f$ is a neural network comprised of a convolutional encoder (top in gray) and an MLP decoder (bottom in gray). The decoder predicts SDF values for 3D points (red/blue circles) through their spatially sampled features (squares in shades of red/blue) from the encoder's activation maps. Following a gradient-based few-shot learning algorithm (MAML \cite{finn2017model}), we learn a meta-decoder in encoder feature space, parameterized with $\boldsymbol{\theta}$, that can quickly adapt to a new shape, \ie new parameters $\boldsymbol{\phi_i}$, given its support. This is achieved by iterating per-shape 5-step adaptation gradient descent (orange arrow) using the support loss $\mathcal{L}_{\mathbf{S}_i}$, and one-step meta gradient-descent (green arrow) by back-propagating the Query set ($\mathbf{Q}_i$) loss $\mathcal{L}_{\mathbf{Q}_i}$ evaluated with the specialized parameters $\boldsymbol{\phi_i}$ \wrt the meta-parameters $\boldsymbol{\theta}$. At test time, 5 fine-tuning iterations are performed similarly starting from the converged meta-model to evaluate $f$.}
\vspace{-10pt}  
  \label{fig:pipe}
\end{figure*}

\noindent\textbf{Reconstruction From a Point Cloud}\quad
Among classical solutions to this task, combinatorial approaches define shapes with a space partitioning based on the input points, using \eg alpha shapes \cite{bernardini1999ball} Voronoi diagrams \cite{amenta2001power} or triangulation \cite{cazals2006delaunay,liu2020meshing,rakotosaona2021differentiable}. Alternatively, implicit function based approaches use the point samples to define a function whose zero level set approximates the surface, through  fitting \eg radial basis functions \cite{carr2001reconstruction}, Gaussian kernels \cite{scholkopf2004kernel}, piece-wise polynomials \cite{ohtake2005sparse},  moving least-squares \cite{kolluri2008provably,liu2021deep}, or by solving a Poisson equation \cite{kazhdan2013screened}. Closer to our scope, recent work proposes to obtain these implicit functions through deep learning. These include two families of work: supervised and unsupervised ones. 

For the latter, a neural network is fitted to the raw input point cloud without any further supervision. Among contributions in this area, Gropp \etal \cite{gropp2020implicit} introduces a regularization on the function's spatial gradient based on the Eikonal equation. Atzmon \etal learns a signed distance function from unsigned distance supervision \cite{atzmon2020sal}, and further supervises the spatial gradient of the function using point normals \cite{atzmon2020sald}. Ma \etal \cite{ma2020neural} supervises the training through expressing the nearest neighbor on the surface as a function of the neural signed distance and its spatial gradient. All of the aforementioned methods benefit from efficient gradient computation through back-propagation in the implicit neural function. \cite{sitzmann2020implicit} introduces periodic activations. \cite{williams2021neural} proposes to learn infinitely wide shallow ReLU networks as random feature kernels. Lipman \cite{lipman2021phase} formalizes a loss ensuring the function converges to occupancy while its log transform converges to a distance function.  

Supervised methods on the other hand assume a training dataset of shapes with ground-truth signed distance or occupancy values for dense space samples \ie points. Auto-decoding based methods \eg \cite{park2019deepsdf,tretschk2020patchnets,jiang2020local,chabra2020deep} require test time optimization to fit the implicit function's features to the observed point cloud, which can take several seconds for a simple object. Conversely, encoder-decoder based approaches enable faster single forward pass inference and superior generalization. 
For these approaches, pooling-based set encoders (\eg PointNet \cite{qi2017pointnet}) were first proposed \cite{mescheder2019occupancy,chen2019learning,genova2020local,erler2020points2surf}, but they have been shown to underfit for large and detailed inputs. More recently, convolutional encoders \cite{chibane2020implicit,peng2020convolutional,liu2021deep}  enable access to more expressive local point features and incorporate inductive biases such as translational equivariance, thus enabling fine-grained implicit reconstruction. 
We propose here to extend such supervised convolutional encoder-decoder models to a few-shot setting to further improve their reconstruction abilities from a sparse point cloud input, while still offering fast inference unlike auto-decoding.


\section{Method}

The task at hand is to recover a continuous shape surface $\mathcal{S}$ given an input point could $\mathbf{X}\subset{\mathbb R}^{3\times N_p}$ representing that underlying shape \ie $\mathbf{X}=\{x_i\sim\mathcal{S} \}_{i=1}^{N_p}$. To this end, we train a deep neural network $f$ to approximate the signed distance function of the target shape given point cloud $\mathbf{X}$. The inferred shape can then be obtained as the zero level set of $f$:
\begin{equation}
\hat{\mathcal{S}} = \{ x\in\mathbb{R}^3 \mid f(\mathbf{X},x) = 0\}.
\end{equation}
We can reconstruct an explicit triangle mesh for shape $\hat{\mathcal{S}}$ using \eg Marching Cubes \cite{lorensen1987marching}. We assume shapes to be watertight manifolds, and that they are all normalized into a domain $\Omega \subset {\mathbb R}^{3}$.

\subsection{The Base Model}
\label{sec:base}

For our neural network $f$, we use an encoder-decoder architecture that follows the model introduced by Chibane \etal \cite{chibane2020implicit}. Such models (\eg \cite{chibane2020implicit,peng2020convolutional}) combining local features extracted with convolutional encoders with implicit decoders have been shown to yield superior performances in the class of single forward pass prediction methods for surface reconstruction. 
Differently from \cite{chibane2020implicit}, we note that we learn signed distance instead of occupancy functions. 

As illustrated in Fig.\ref{fig:pipe}, The encoder takes as input point cloud $\mathbf{X}$ and produces spatial feature maps. In order to apply this 3D convolutional network to the point cloud, the latter is first voxelized into a discrete 3D grid in $\mathbb{R}^{N\times N\times N}$ (\cf Fig.\ref{fig:vox}), $N$ being the input spatial resolution. It then passes through successive convolutional down-sampling blocks resulting in $n$ multi-scale deep feature grids $\mathbf{F}_1,\dots,\mathbf{F}_n$, where $\mathbf{F}_k\in\mathbb{R}^{C_k\times N_k \times N_k \times N_k}$. The feature map channels $C_k$ increase with the encoder's depth while their resolution decreases $N_k=N/{2^{k-1}}$. The shallow features represent local details while the deeper ones account for more global shape variation. Given a 3D point $x\in\Omega$, we can extract its encoder generated features using trilinear interpolation. We define this process with a neural function $\Psi_{\mathbf{X}}:\Omega\rightarrow\mathbb{R}^{C_1}\times\dots\times\mathbb{R}^{C_n}$ such that:
\begin{equation}
\Psi_{\mathbf{X}}(x)=(\mathbf{F}_1(x),\dots,\mathbf{F}_n(x)).  
\end{equation}

The decoder is tasked with predicting the signed distance to the ground-truth shape $\mathcal{S}$ for a given 3D point $x$. It uses the features obtained with the encoder as input point representation. It consists of a MLP with ReLU non-linearities and a final Tanh activation, and we denote it as $\Phi:\mathbb{R}^{C_1}\times\dots\times\mathbb{R}^{C_n}\rightarrow\mathbb{R}$. Hence we can express the approximated signed distance function given a point cloud $\mathbf{X}$ as follows:
\begin{equation}
f(\mathbf{X},x) = \Phi \circ \Psi_\mathbf{X}(x).
\end{equation}

In standard supervised learning, this network is trained by back-propagating the prediction loss over a set of training points $\mathbf{Y}\subset\Omega$ per training shape using their respective pre-computed ground-truth signed distance values. These dense point sets are typically built by sampling near the ground-truth surface $\mathcal{S}$, \ie sampling points on the surface and offsetting them with normally distributed displacements:
\begin{equation}
\mathbf{Y}=\{x+n:x\sim\mathcal{S},n\sim\mathcal{N}(0,\mathbf{\Sigma})\},
\label{equ:samp}
\end{equation}
where $\mathbf{\Sigma}=\text{diag}(\sigma) \in \mathbb{R}^{3\times 3}$ is a diagonal covariance matrix. An illustrative example of such a set can be seen in the bottom left of Fig.\ref{fig:pipe}.

\subsection{Few-shot Learning in Feature Space}

We would like to build a model $f$ that can learn to adapt to a new shape $\mathcal{S}$ given limited observations, namely the input point cloud $\mathbf{X}$. While the network is already conditioned to the input $\mathbf{X}$ through the encoder in the standard supervised learning regime \eg \cite{chibane2020implicit,peng2020convolutional}, we seek here to adapt it even further to that input through meta-learning. Let us recall that for each training shape $\mathcal{S}_i$ we have two sets of points available: $\mathbf{X}_i$ the sparse input point cloud at the surface, and $\mathbf{Y}_i$ the dense point set sampled near the ground-truth surface. Corresponding ground-truth signed distances are available for both of these sets as well. However at test time, only $\mathbf{X}_i$ is available.\\  

\vspace{-5pt}
\noindent\textbf{Support and Query Sets}\quad
We position ourselves in a meta-learning based few-shot learning setup \cite{vinyals2016matching,snell2017prototypical,finn2017model}. Traditionally, a network is trained to adapt to a new task given limited training samples in this setup. A task is defined with a loss, a support (or context) set and a query set. These sets are input-target pairs for the given task. The model is trained to perform tasks on their query sets, after being adapted to them through \eg metric learning \cite{vinyals2016matching} or gradient descent \cite{finn2017model} on their respective limited support sets. We adopt the same strategy, where a task consists in learning the signed distance function $f$ for a given shape $\mathcal{S}_i$. We define the support set as the pairs made of the points of the input point cloud and their corresponding ground-truth signed distances, as such a set is available at test time:
\begin{align}
\mathbf{S}_i & =\{(x,s):x\in\mathbf{X}_i,s:=\text{SDF}(x)\}.\\
 & =\{(x,0):x\in\mathbf{X}_i\}.
\end{align}
Since $\mathbf{X}_i$ contains exclusively points from the surface, \ie the zero level set of the shape function, all ground-truth singed distances are null. 
We define the query set as the pairs made of the dense points pre-sampled around the surface and their corresponding ground-truth singed distance values:
\begin{equation}
\mathbf{Q}_i=\{(x,s):x\in\mathbf{Y}_i,s:=\text{SDF}(x)\}.
\end{equation}\\

\begin{algorithm}
\scriptsize
\begin{algorithmic}
\Require Dataset fo shapes $\mathcal{S}_i$, pre-trained encoder $\Psi$, meta-decoder learning rate $\beta$
\Ensure meta-decoder weights $\boldsymbol{\theta}$, decoder learning rates $\boldsymbol{\alpha}$
\State initialize $\boldsymbol{\theta}$, $\boldsymbol{\alpha}$ 
\While{$\text{not done}$}
\State sample batch of shapes $\{\mathcal{S}_i\}:=\{(\mathbf{X}_i,\mathbf{Q}_i)\}$
\State initialize $\mathcal{L}_{\mathbf{Q}}\leftarrow0$  
\For{$\mathcal{S}_i$ in $\{\mathcal{S}_i\}$}
\State initialize $\boldsymbol{\phi}_i\leftarrow\boldsymbol{\theta}$
\For{$K$ \text{times}}
\State $\mathcal{L}_{\mathbf{S}_i} = \sum_{x\in\mathbf{X}_i}|\Phi_{\boldsymbol{\phi}_i} \circ \Psi_{\mathbf{X}_i}(x) |$
\State $\boldsymbol{\phi}_i \leftarrow \boldsymbol{\phi}_i-\boldsymbol{\alpha} \odot \nabla_{\boldsymbol{\phi}_i}\mathcal{L}_{\mathbf{S}_i}$ 
\EndFor
\State $\mathcal{L}_{\mathbf{Q}} \leftarrow \mathcal{L}_{\mathbf{Q}} + \sum_{(x,s)\in \mathbf{Q}_i}|\Phi_{\boldsymbol{\phi}_i} \circ \Psi_{\mathbf{X}_i}(x)-s|$
\EndFor
\State $(\boldsymbol{\theta},\boldsymbol{\alpha}) \leftarrow (\boldsymbol{\theta},\boldsymbol{\alpha}) - \beta \nabla_{\boldsymbol{\theta},\boldsymbol{\alpha}} \mathcal{L}_{Q}$
\EndWhile
\end{algorithmic}
\caption{\small The training procedure of our model.}
\label{alg:train}
\end{algorithm}

\vspace{-20pt}
\noindent\textbf{Meta-Learning in Feature Space}\quad
We apply gradient-based meta-learning to our supervised few-shot shape function learning, in particular the MAML algorithm by Finn \etal \cite{finn2017model}. For a given shape $\mathcal{S}_i$, and assuming a pre-trained encoder $\Psi$, the signed distance function $f$ is obtained through a specialization denoted $\boldsymbol{\phi}_i$ of the parameters $\boldsymbol{\theta}$ of an underlying meta-decoder $\Phi_{\boldsymbol{\theta}}$ operating in feature space $\Psi_{\mathbf{X}_i}(\Omega)$: 
\begin{equation}
f(\mathbf{X}_i,x) = \Phi_{\boldsymbol{\phi}_i} \circ \Psi_{\mathbf{X}_i}(x).     
\end{equation}
For lower computational and memory costs and a less noisy meta-learning loss (\cf Section \ref{sec:single}), we fix the convolutional encoder $\Psi$ after pre-training it. This encoder is pre-trained by training the base model $f$ in the standard supervised learning regime using the training dataset's query sets $\{\mathbf{Q}_i\}$ for supervision (\i.e. standard supervised learning). As such, the meta-learning of model $f$ consists in training the meta-decoder $\Phi_{\boldsymbol{\theta}}$ in feature space. Each training step in this process is two fold: First, a fixed number of inner training steps, \ie adaptation of the meta-decoder $\Phi_{\boldsymbol{\theta}}$ into $\Phi_{\boldsymbol{\phi}_i}$, followed by an outer training step, \ie update of the meta-decoder $\Phi_{\boldsymbol{\theta}}$. Similarly to Sitzmann \etal \cite{sitzmann2020metasdf}, we build on the Meta-SGD \cite{li2017meta} MAML \cite{finn2017model} variant proposed by Li \etal, which advocates the use of per-parameter learning rates in the adaptation stage for improved flexibility. 

Given a batch of training shapes $\{\mathcal{S}_i\}$, the inner training step of the decoder is performed for each shape $\mathcal{S}_i$ independently. The $L_1$ loss $\mathcal{L}_{\mathbf{S}_i}$ is computed using the current specialized decoder $\Phi_{\boldsymbol{\phi}_i}$ over the support set $\mathbf{S}_i$ (\ie $\mathbf{X}_i$), and is back-propagated \wrt $\boldsymbol{\phi}_i$:
\begin{gather}
\mathcal{L}_{\mathbf{S}_i} = \sum_{x\in\mathbf{X}_i}|\Phi_{\boldsymbol{\phi}_i} \circ \Psi_{\mathbf{X}_i}(x) |,\\
\boldsymbol{\phi}_i \leftarrow \boldsymbol{\phi}_i - \boldsymbol{\alpha} \odot \nabla_{\boldsymbol{\phi}_i}\mathcal{L}_{\mathbf{S}_i},
\end{gather}
where weights $\boldsymbol{\phi}_i$ are initialized with the current meta-decoder weights $\boldsymbol{\theta}$ for all the batch shapes. $\boldsymbol{\alpha}$ contains the per parameter learning rates, which are learned as part of the outer training loop. $\odot$ symbolizes element-wise product. We note that while the support loss $\mathcal{L}_{\mathbf{S}_i}$ could include additional regularisation such as the Eikonal constraint \cite{gropp2020implicit}, we keep it simple to limit the computational footprint of the meta-learning.

After $K$ such shape specific adaptation steps, one outer training step is performed for the entire batch of shapes. The $L_1$ losses $\{\mathcal{L}_{\mathbf{Q}_i}\}$ are computed using the specialized decoders $\{\Phi_{\boldsymbol{\phi}_i}\}$ over their respective query sets $\{\mathbf{Q}_i\}$, and their average is back-propagated \wrt the meta-parameters $\boldsymbol{\theta}$ and $\boldsymbol{\alpha}$ accordingly:
\begin{gather}
\mathcal{L}_{\mathbf{Q}_i} = \sum_{(x,s)\in \mathbf{Q}_i}|\Phi_{\boldsymbol{\phi}_i} \circ \Psi_{\mathbf{X}_i}(x)-s|,\\ 
(\boldsymbol{\theta},\boldsymbol{\alpha}) \leftarrow (\boldsymbol{\theta},\boldsymbol{\alpha}) - \beta \nabla_{\boldsymbol{\theta},\boldsymbol{\alpha}} \sum_i \mathcal{L}_{\mathbf{Q}_i},
\end{gather}
where $\beta$ is a scalar learning rate. For ease of understanding, Algorithm \ref{alg:train} provides a summary of this training procedure.

At test time, given an input $\mathbf{X}$, the inference consists in a forward pass of the model $f$ after a 
$K$-step adaptation of the converged meta-decoder $\Phi_{\boldsymbol{\theta}}$. To produce mesh reconstructions, we use the model to predict signed distance values of a grid of points at a desired resolution, and then apply the Marching Cubes \cite{lorensen1987marching} algorithm on the inferred signed distance grid. 


\section{Results}
\label{sec:res}

\begin{figure}[t!]
\centering
\includegraphics[width=0.5\linewidth]{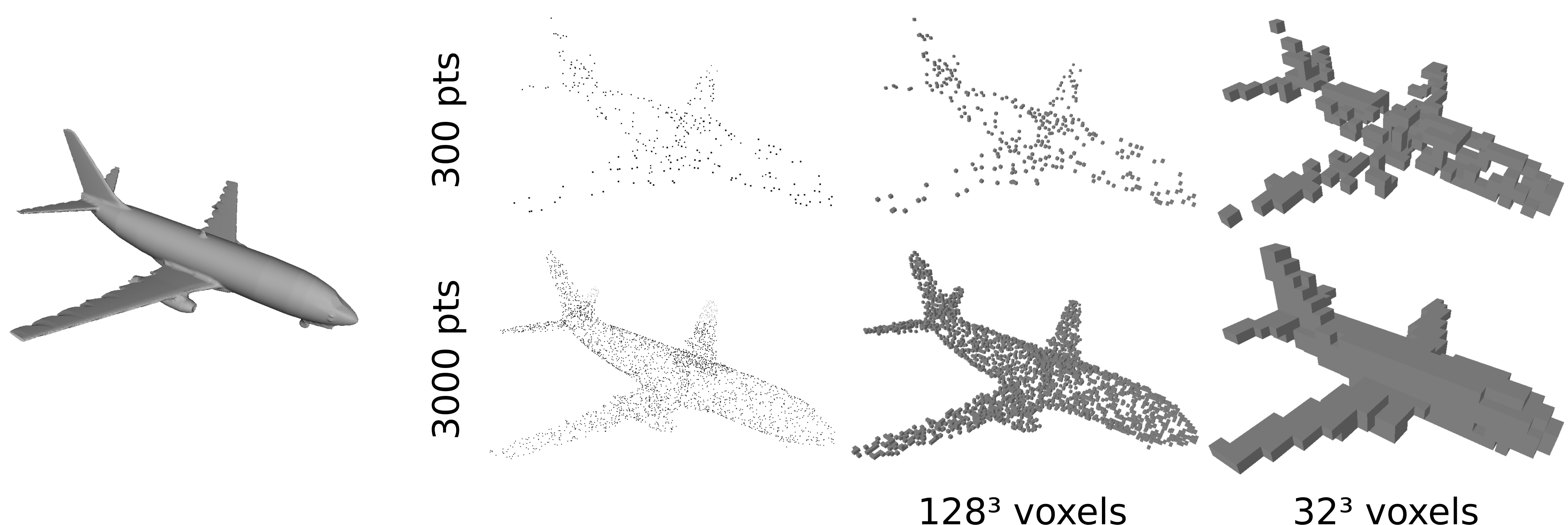}
\vspace{-2pt} 
  \caption{Visualization of voxelizations at resolutions $32^3$ and $128^3$ of input point clouds with $300$ and $3000$ points.}
  \label{fig:vox}
\vspace{-10pt}  
\end{figure}

We present in this section our experimental setup and showcase our results. We evaluate our method on both multi-class and single class setups in ShapeNet \cite{shapenet} on reconstruction from a sparse point cloud, and we also show results on the FAUST \cite{Bogo:CVPR:2014} dataset. We follow the noise-free benchmark in our baselines IF-Nets \cite{chibane2020implicit} and MetaSDF \cite{sitzmann2020metasdf}. We experiment with two sizes of input point clouds $N_p=3000$ and $N_p=300$ similarly to \cite{chibane2020implicit}, and two voxelization resolutions $N=128$ and $N=32$ (Fig.\ref{fig:vox}) of these point sets. Let us recall that inputs require voxelization due to the 3D convolutional encoder of our base model \cite{chibane2020implicit}. In summary, our proposed approach outperforms all baselines including the same base model trained in standard supervised learning (IF-Nets \cite{chibane2020implicit}), and the decoder of the base model trained through meta-learning (MetaSDF \cite{sitzmann2020metasdf}). Results show additionally that our approach is more resilient to coarser inputs compared to IF-Nets. Besides, the performance increase brought by our approach comes with minimal additional computational cost, as inference takes 150 ms for our model, and 60 ms for IF-Nets, on a RTX A4000.

\subsection{Implementation Details}

The base model follows the architecture in \cite{chibane2020implicit}, hence we use $n=6$ feature maps with feature dimensions $C_1=1$, $C_2=16$, $C_3=32$, $C_4=64$, $C_5=128$ and $C_6=128$. Regarding the meta-learning, we use $K=5$ steps in the inner training loop and we initialize the per-parameter learning rates $\alpha$ with $10^{-6}$. In the outer loop, we set the meta-decoder learning rate to $\beta=10^{-6}$. We train for $100$ epochs with batches of $4$ shapes, leveraging the $N_p$ training points in the inner loop, and $50$k training points in the outer loop per shape. To train the base model in the standard supervised learning mode, we perform a maximum of $50$ epochs with a learning rate of $10^{-5}$, using batches of $8$ shapes with $50$k training points per shape. All trainings use the Adam\cite{KingmaB14} solver on a RTX A4000 in the PyTorch \cite{paszke2019pytorch} framework. All Marching Cubes reconstructions are done with a $256^3$ sized grid. 

\subsection{Datasets}

Similar to prior work we evaluate our method using the ShapeNet benchmark \cite{shapenet} which consists of various instances of $13$ different object classes. Similarly to \cite{chibane2020implicit}, we use the pre-processing by \cite{xu2019disn} to obtain watertight meshes which enables computing ground-truth signed distances. All meshes are subsequently normalized using their bounding boxes thus fitting inside the domain $\Omega=[-1,1]^3$. We use the train/test split provided by \cite{chibane2020implicit}, which is based on the original split of Choy \etal \cite{choy20163d} minus 508 distorted shapes due to pre-processing failures. To create the input point cloud $\mathbf{X}$ for a given shape, $N_p$ sized sets of points are randomly pre-sampled from the processed mesh. For the training points with ground-truth signed distances, we pre-sample $100$k points near the surface with $\sigma = 0.1$ and $\sigma = 0.01$ (\cf Equ. \ref{equ:samp}). At training, $50$k points are sampled equally from these pre-made two sets to make the per shape training points batch $\mathbf{Y}$. We also use the FAUST dataset \cite{Bogo:CVPR:2014} for testing. It consists of 100 registered meshes of 10 human body identities in 10 different poses.

\begin{figure}[t!]
\centering
\includegraphics[width=0.6\linewidth]{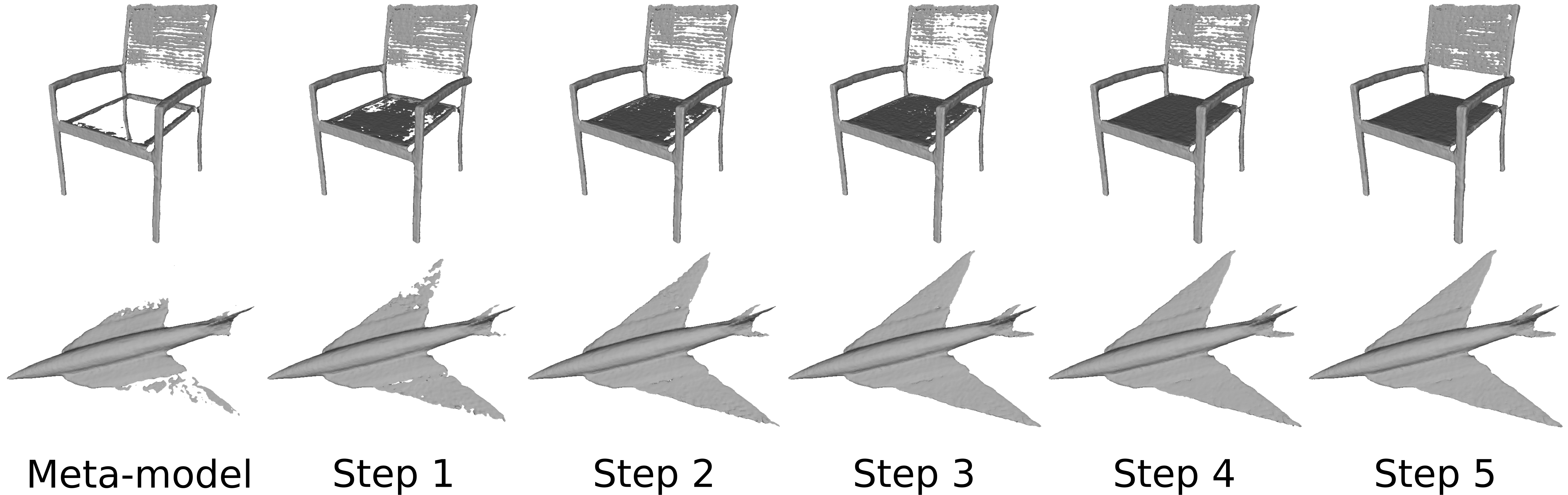}
  \caption{Visualization of reconstructions from $3000$ points throughout the $K=5$ testing inner-loop iterations in the ShapeNet multi-class setup.}
  \label{fig:steps}
\vspace{-7pt}  
\end{figure}

\subsection{Metrics}

We evaluate our method and baselines using popular metrics for 3D reconstruction quality assessment. We denote here by $\mathcal{S}$ and $\hat{\mathcal{S}}$ respectively the ground-truth and the generated shape meshes. The volumetric Intersection over Union (IoU) is defined as the ratio of the the intersection of the inside volumes of the meshes to their union, and it is implemented following \cite{mescheder2019occupancy} accordingly:
\begin{equation}
\small
\text{IoU} = \frac{|\{x\in\Omega:x\, \text{inside}\, \mathcal{S}\, \text{and}\, \hat{\mathcal{S}}\}|}{|\{x\in\Omega:x\,\text{inside}\,\mathcal{S}\,\text{or}\,\hat{\mathcal{S}}\}|},
\end{equation}
where $|.|$ symbolizes the cardinality of the sets, which is approximated by sampling 100k points in the bounding volume $\Omega$. We also report two variants of the Chamfer distance representing the two-ways nearest neighbor distance between the meshes, using averaged minimal distances for $\text{CD}_\text{1}$ \cite{mescheder2019occupancy} and averaged minimal squared distances for $\text{CD}_\text{2}$ \cite{park2019deepsdf}:
{\small
\begin{gather}
\small
\text{CD}_\text{1} = \frac{1}{2|\mathcal{S}|}\sum_{v\in\mathcal{S}}\min_{\hat{v}\in\hat{\mathcal{S}}}\Vert v-\hat{v} \Vert_2 +
\frac{1}{2|\hat{\mathcal{S}}|}\sum_{\hat{v}\in\hat{\mathcal{S}}}\min_{v\in\mathcal{S}}\lVert \hat{v}-v \rVert_2,\\
\text{CD}_\text{2} = \frac{1}{2|\mathcal{S}|}\sum_{v\in\mathcal{S}}\min_{\hat{v}\in\hat{\mathcal{S}}}\Vert v-\hat{v} \Vert_2^2 +
\frac{1}{2|\hat{\mathcal{S}}|}\sum_{\hat{v}\in\hat{\mathcal{S}}}\min_{v\in\mathcal{S}}\Vert \hat{v}-v \Vert_2^2.
\end{gather}
}%
The metrics are also approximated here with $100$k samples from the source and target meshes, where distances are computed using a KD-tree following \cite{chibane2020implicit}. In the remainder on the paper, we report $\text{CD}_\text{2} \times 10^{-3}$ and $\text{CD}_\text{1} \times 10^{-1}$.

\begin{figure*}[t!]
\centering
\includegraphics[width=\linewidth]{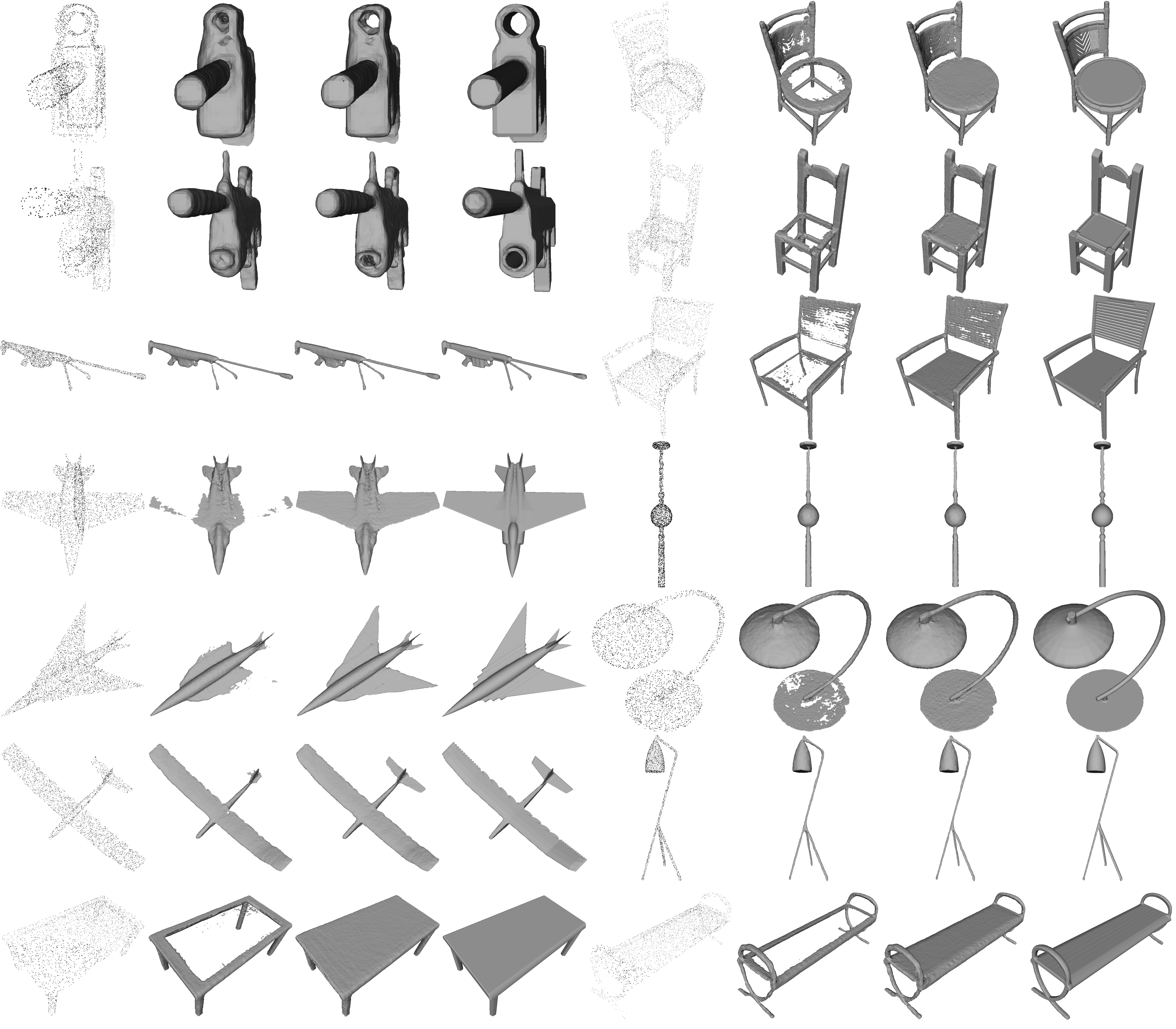}
  \caption{Qualitative comparison of reconstructions from 3000 points on ShapeNet with our main baseline, \ie same base model in standard supervised learning. (Input / IF-Nets\cite{chibane2020implicit} / \textbf{Ours} / Ground-truth).}
  \label{fig:qual3k}
\vspace{-7pt}  
\end{figure*}

\subsection{Multiple Shape Class Evaluation}
\label{sec:multi}

We evaluate here our work and the competition using the entire ShapeNet dataset, which counts 26834 training shapes and 7148 testing ones. For the input point cloud size and voxelization resolution of our method, we consider here the two extreme cases for brevity: \ie 3000 points at 128³ resolution (Tab.\ref{tab:3k-128}), and 300 points at 32³ resolution (Tab.\ref{tab:300-32}). We refer the reader to a more detailed analysis in the ablative single class evaluation (Section \ref{sec:single}). We relay the performances of OccNet \cite{mescheder2019occupancy}, DMC \cite{liao2018deep}, PSGN \cite{fan2017point} as they were reported in \cite{chibane2020implicit}. We reproduced the performance of ConvOccNet \cite{peng2020convolutional} with 3000 input points in the noise-free benchmark of IF-Nets \cite{chibane2020implicit} and we obtained an IoU of 0.86, which is also confirmed by the results of other contemporary work \cite{williams2022neural} (see fig. 9 in that paper). The authors of ConvOccNet report higher numbers (0.88), where noise is added to the input point cloud. For fairness, we report ConvOccNet's higher numbers (\ie 0.88 IoU). We train our own IF-Nets \cite{chibane2020implicit} model on signed distances and reproduce the same results in the main paper. MetaSDF here refers to our implementation of the work in \cite{sitzmann2020metasdf} with 3000 input surface points, \ie our model without an encoder trained for hundreds of epochs, from which we perform numerous evaluations and pick the best one. We note that original paper \cite{sitzmann2020metasdf} only showed results for 10k input points. This same number is reported in table \ref{tab:300-32} under the name MetaSDF (3k pts).

\begin{table}[t!]
\begin{subtable}{.49\linewidth}
\centering
\scalebox{0.8}{\begin{tabular}{|c|c|c|c|}
\cline{2-4}
\multicolumn{1}{c|}{} & IoU\footnotesize{$\uparrow$} & $\text{CD}_\text{1}$\footnotesize{$\downarrow$} & $\text{CD}_\text{2}$\footnotesize{$\downarrow$}\\
\hline
OccNet & 0.72 & -- & 0.4 \\
DMC & 0.65 & -- & 0.1 \\
PSGN & -- & -- & 0.4 \\
\hline
MetaSDF & 0.63 & 0.123 & 0.458 \\
ConvOccNet & 0.88 & 0.044 & -- \\
IF-Nets & 0.88 & 0.032 & 0.032 \\
\textbf{Ours} & \textbf{0.91} & \textbf{0.028} & \textbf{0.026} \\
\hline
\end{tabular}}
\vspace{-5pt}
\caption{}
\label{tab:3k-128}
\end{subtable}
\begin{subtable}{.49\linewidth}
\scalebox{0.8}{\begin{tabular}{|c|c|c|c|}
\cline{2-4}
\multicolumn{1}{c|}{} & IoU\footnotesize{$\uparrow$} & $\text{CD}_\text{1}$\footnotesize{$\downarrow$} & $\text{CD}_\text{2}$\footnotesize{$\downarrow$}\\
\hline
OccNet & 0.73 & -- & 0.3 \\
DMC & 0.58 & -- & 0.3 \\
PSGN & -- & -- & 0.4 \\
\hline
MetaSDF (3k pts) & 0.63 & 0.123 & 0.458 \\
IF-Nets & 0.67 & 0.091 & 0.232 \\
\textbf{Ours} & \textbf{0.74} & \textbf{0.070} & \textbf{0.209} \\
\hline
\end{tabular}}
\vspace{-5pt}
\caption{}
\label{tab:300-32}
\end{subtable}
\vspace{-10pt}
\caption{Reconstruction on ShapeNet from (a) 3000 points voxelized at resolution 128³, and (b) 300 voxelized at 32³.}
\vspace{-10pt}
\end{table}



Tables \ref{tab:3k-128} and \ref{tab:300-32} report the average reconstruction performance from 3000 and 300 input points respectively on the entire multi-class testing set. We additionally provide the per-class numbers in the supplementary material. PSGN generates point sets with competitive distances to the ground-truth but does not provide any connectivity (Hence the absence of IoU). DMC's performance is limited by its voxel grid resolution. OccNet performs strongly and almost similarly in the 300 and 3000 input cases, which suggests that pooling set encoders underfit the context. 
For both input situations and across all metrics, our method outperforms the competition, including convolutional encoder equipped implicit shape models (IF-Nets and ConvOccNet), and our encoder-free meta-learning baseline MetaSDF. We find the performance of the latter particularly underwhelming, which suggests that despite the encouraging single class reconstruction results in \cite{sitzmann2020metasdf} from 10k input surface points, such strategy struggles to scale to more challenging settings with multiple classes of shape, sparser point clouds, and under relatively limited training time. When decreasing the input size and voxel resolution in Table \ref{tab:300-32}, both the encoders of our method and IF-Nets are exposed to very poor inputs (\cf Fig \ref{fig:vox}).
While the performance of IF-Nets is heavily affected by these coarser inputs, our method is more resilient thanks to the meta-learning addition. In fact, our IoU drops by 18\% compared to 23\% for the standard supervisedly learned baseline. We additionally show reconstruction results on the FAUST dataset from models trained on ShapeNet in Table \ref{tab:faust}, where out method outperforms IF-Nets. Note that neither models have seen human shapes nor articulated shapes for that matter in training. 



\begin{table}[t!]
    \centering
\scalebox{0.8}{
\begin{tabular}[b]{|c|c|c|c|}
\cline{2-4}
\multicolumn{1}{c|}{} & IoU\footnotesize{$\uparrow$} & $\text{CD}_\text{1}$\footnotesize{$\downarrow$} & $\text{CD}_\text{2}$\footnotesize{$\downarrow$}\\
\hline
IF-Nets & 0.82 & 0.037 & 0.060 \\
\textbf{Ours} & \textbf{0.84} & \textbf{0.035} & \textbf{0.051} \\
\hline
\end{tabular}
\qquad
\includegraphics[width=0.7\textwidth]{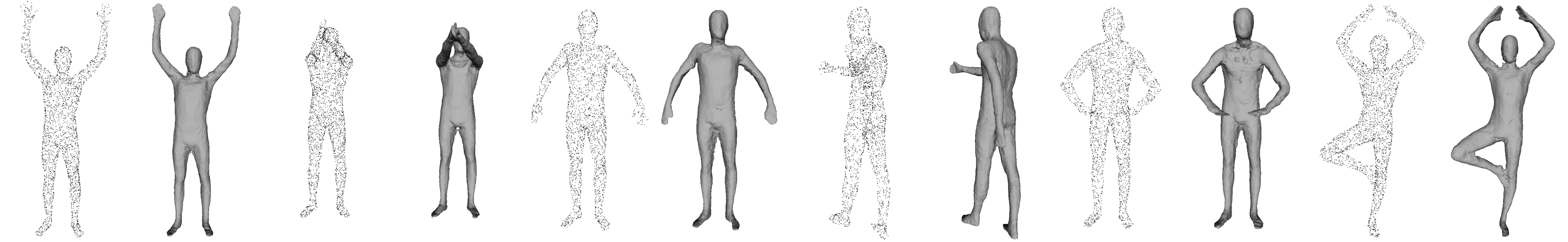}    
}
    \vspace{1pt}
    \caption{Reconstruction on FAUST with models trained on ShapeNet, from 3000 points voxelized at resolution 128³. Qualitative results from our model.}
    \label{tab:faust}
\vspace{-15pt}    
\end{table}

The numerical superiority of our method \wrt our closest baseline (IF-Nets) is supported with qualitative comparisons in Fig.\ref{fig:qual3k}. We notice that our method manages to recover more thin structures and fine topological features, such as cylindrical holes in rifles, wings in planes, thin flat surfaces in tables, benches and chairs. In addition, we provide examples in Fig.\ref{fig:steps} showing the evolution of the reconstruction at various iterations of our inference.


\subsection{Single Shape Class Ablation}
\label{sec:single}

We show further quantitative evaluations in a single shape class setup on the largest class of ShapeNet, \ie table, for ablative purposes and also for a more fair comparison to MetaSDF \cite{sitzmann2020metasdf}. The table class counts 5364 training shapes and 1679 testing ones. We report numbers for MetaSDF and a pooling set encoder based implicit model (PoinNet enc.) from their paper \cite{sitzmann2020metasdf} using 10k input points. We compare multiple variants of our approach. For our method (Ours), the meta-decoder is initialized from the weights of a pre-trained base model. For Ours w/o dec. pret., it is initiated with the standard PyTorch initialization instead. 
Finally  Ours w/o meta learn. (\ie IF-Nets) is again our base model trained in standard supervised learning.

Tables \ref{tab:128-table} and \ref{tab:32-table} show reconstruction results from 3000/300 points at 128³ and 32³ input voxel resolution respectively. 
Even under 10k input points, MetaSDF \cite{sitzmann2020metasdf} can yet barely reach our performance on just 300 input points. 
Most notably, our method improves on the standard supervised learning baseline (Ours w/o meta learn. (\ie IF-Nets)) across all input sizes and input voxel discretizations for all metrics. While the input point cloud size affects both our method and IF-Nets almost equally, decreasing the encoder's input resolution hinders IF-Nets's performance substantially more severely. In fact, when going from 128³ to 32³ resolution inputs, our IoU drops by roughly 14\% vs. 20\% for IF-Nets for 3000 points, and by 13\% vs. 19\% for IF-Nets when using 300 points.\\

\vspace{-7pt}
\noindent\textbf{Decoder pre-training}\quad
As witnessed by tables \ref{tab:128-table} and \ref{tab:32-table}, while initializing the meta-decoder can improve the performance slightly, it is not crucial for obtaining satisfactory results, which suggests that we learn a proper meta-decoder and not just a fine-tuned base-model.\\

\vspace{-7pt}
\noindent\textbf{Encoder pre-training and tuning}\quad
As meta-learning both the encoder and decoder is computationally and memory expensive, we only meta-learn the decoder. We found that tuning the encoder during this meta-learning leads to noisy losses, without a clear improvement in the results. In fact, for reconstruction from 3000 points at resolution $128^3$ in class table, whilst fixing the encoder yields an IoU of $0.87$, tuning it gives a comparable performance ($0.86$) while requiring more time and memory for training. We found the resulting noisier loss makes it also harder to decide the convergence epoch in this case. Thus we fix the encoder after pre-training it. We pre-train the encoder by training the encoder-decoder in the standard supervised learning setup.

\begin{table}[t!]
    \centering
\scalebox{0.8}{\begin{tabular}{|c|cc|cc|cc|}
\cline{2-7}
\multicolumn{1}{c|}{} & \multicolumn{2}{c|}{IoU\footnotesize{$\uparrow$}} & \multicolumn{2}{c|}{$\text{CD}_\text{1}$\footnotesize{$\downarrow$}} & \multicolumn{2}{c|}{$\text{CD}_\text{2}$\footnotesize{$\downarrow$}}\\
\hline
PointNet enc. (10k pts) & \multicolumn{2}{c|}{0.66}  & \multicolumn{2}{c|}{--} & \multicolumn{2}{c|}{0.69} \\
MetaSDF (10k pts) & \multicolumn{2}{c|}{0.75} & \multicolumn{2}{c|}{--} & \multicolumn{2}{c|}{0.32} \\
\hline 
Ours w/o meta learn. (\ie IF-Nets) &  0.82 & 0.72 & 0.040 & 0.057 & 0.062 & 0.097 \\
Ours w/o dec. pret. & 0.86 & 0.74 & 0.035 & 0.057 & 0.030 & 0.203 \\
\textbf{Ours} & \textbf{0.87} &\textbf{ 0.76} &\textbf{ 0.033}  & \textbf{0.051} & \textbf{0.030}  &\textbf{0.082} \\
\hline
    \end{tabular}}
    \vspace{1pt}
    \caption{Reconstruction on class table of ShapeNet from 3000 (left) and 300 (right) points voxelized at resolution 128³.}
    \label{tab:128-table}
\vspace{-10pt}    
\end{table}

\begin{table}[t!]
    \centering
\scalebox{0.8}{\begin{tabular}{|c|cc|cc|cc|}
\cline{2-7}
\multicolumn{1}{c|}{} & \multicolumn{2}{c|}{IoU\footnotesize{$\uparrow$}} & \multicolumn{2}{c|}{$\text{CD}_\text{1}$\footnotesize{$\downarrow$}} & \multicolumn{2}{c|}{$\text{CD}_\text{2}$\footnotesize{$\downarrow$}}\\
\hline
Ours w/o meta learn. (\ie IF-Nets) & 0.65 & 0.58 & 0.071 & 0.092 & 0.089 & 0.169 \\
Ours w/o dec. pret. & 0.73 & 0.61 & 0.057 & 0.083 & 0.082  & 0.169  \\
\textbf{Ours} &\textbf{0.74} &\textbf{0.66} & \textbf{0.052} & \textbf{0.076} &\textbf{0.068}  &\textbf{0.142}  \\
\hline
    \end{tabular}}
    \vspace{1pt}
    \caption{Reconstruction on class table of ShapeNet from 3000 (left) and 300 (right) points voxelized at resolution 32³.}
    \label{tab:32-table}
\vspace{-10pt}
\end{table}
\vspace{-10pt}
\section{Limitations}
\vspace{-10pt}
As SDFs can only represent closed surfaces, we will experiment next with other representations such as points \cite{venkatesh2021deep,cai2020learning} and unsigned distances \cite{chibane2020neural}. Point cloud voxelization (\cf Fig.\ref{fig:vox}) hinders the expressiveness of the input, thus we will be considering different convolutional encoders subsequently. Furthermore, the MAML algorithm \cite{finn2017model} requires computing second-order gradients which raises the memory complexity in training. Finally, we follow here the noise-free benchmarks in our baselines IF-Nets \cite{chibane2020implicit} and MetaSDF \cite{sitzmann2020metasdf}. Considering noisy and real inputs (\eg 2.5D, SFM, etc.) is part of our future work.

\vspace{-10pt}
\section{Conclusion}
\vspace{-10pt}
We proposed to perform 3D shape reconstruction from a sparse point cloud using a implicit neural model conditioned with both encoder generated local features and meta-learning simultaneously. Our results demonstrate numerically and qualitatively that this approach improves on its standard supervised learning counterpart with minimal additional test time computational cost, and this performance gap increases for coarser inputs. Future avenues of improvement include tackling more real world downstream tasks such as partial shape reconstruction, making use of normals, meta-learning of reconstruction from images and depth maps through differentiable rendering, and exploring other meta-learning techniques.  



{\small
\bibliographystyle{splncs04}
\bibliography{egbib}
}

\clearpage

\section*{Supplementary Material}

\subsection*{Noisy input point cloud}

We show in Table \ref{tab:128-table-noise} that when dealing with noisy input point clouds (\eg variance of 0.005), our approach (\ie meta-learning in feature space) still outperforms standard supervised learning (\eg IF-Nets \cite{chibane2020implicit}). We report in this experiment reconstruction results from training and testing on the largest class of ShapeNet \cite{shapenet}: table, after less than a $100$ meta-learning epochs.  

\begin{table}[h!]
    \centering
\scalebox{0.8}{\begin{tabular}{|c|cc|cc|cc|}
\cline{2-7}
\multicolumn{1}{c|}{} & \multicolumn{2}{c|}{IoU\footnotesize{$\uparrow$}} & \multicolumn{2}{c|}{$\text{CD}_\text{1}$\footnotesize{$\downarrow$}} & \multicolumn{2}{c|}{$\text{CD}_\text{2}$\footnotesize{$\downarrow$}}\\
\hline
Ours w/o meta learn. (\ie IF-Nets) &  0.85 & 0.71 & 0.036 & 0.060 & 0.035 & 0.098 \\
\textbf{Ours} & \textbf{0.87} &\textbf{0.74} &\textbf{0.034}  & \textbf{0.056} & \textbf{0.029}  &\textbf{0.089} \\
\hline
    \end{tabular}}
    \vspace{1pt}
    \caption{Reconstruction on class table of ShapeNet from 3000 (left) and 300 (right) noisy points voxelized at resolution 128³.
    ($\text{CD}_\text{1} \times 10^{-1}$, $\text{CD}_\text{2} \times 10^{-3}$).}
    \label{tab:128-table-noise}
\vspace{-7pt}    
\end{table}

\subsection*{Per-class results for tables \ref{tab:3k-128} \& \ref{tab:300-32} in the main submission}

We show next the per-class reconstruction results on the ShapeNet \cite{shapenet} benchmark, from 3000 input points at voxelization resolution 128³, and from 300 at resolution 32³. The average reconstruction scores were reported in tables 1 \& 2 in the main submission. We report IoU in table \ref{tab:iou}, $\text{CD}_\text{2}$ in table \ref{tab:cd2}, and $\text{CD}_\text{1}$ in table \ref{tab:cd1}. We note that IF-Nets\cite{chibane2020implicit} is also our method without meta-learning, and MetaSDF\cite{sitzmann2020metasdf} is also our method without encoder. For MetaSDF, we report the numbers for 3000 input points.     

\begin{table}[h!]
\centering
\scalebox{0.8}{\begin{tabular}{|c|C{45pt}|C{25pt}C{25pt}|C{25pt}C{25pt}|}
\cline{2-6}
\multicolumn{1}{c|}{} & MetaSDF \footnotesize{(3k pts)} & \multicolumn{2}{c|}{\multirow{2}{*}{IF-Nets}} &  \multicolumn{2}{c|}{\multirow{2}{*}{\textbf{Ours}}}\\
\hline
airplane   & 0.64 & 0.71 & 0.90 & \textbf{0.78} & \textbf{0.92}\\
bench      & 0.44 & 0.44 & 0.82 & \textbf{0.59} & \textbf{0.86}\\
cabinet    & 0.65 & 0.66 & 0.81 & \textbf{0.70} & \textbf{0.82}\\
car        & 0.77 & 0.78 & 0.91 & \textbf{0.81} & \textbf{0.91}\\
chair      & 0.55 & 0.63 & 0.88 & \textbf{0.71} & \textbf{0.90}\\
display    & 0.66 & 0.69 & 0.92 & \textbf{0.81} & \textbf{0.95}\\
lamp       & 0.40 & 0.52 & 0.83 & \textbf{0.54} & \textbf{0.85}\\
phone      & 0.83 & 0.79 & 0.94 & \textbf{0.90} & \textbf{0.97}\\
rifle      & 0.58 & 0.63 & 0.87 & \textbf{0.72} & \textbf{0.90}\\
sofa       & 0.79 & 0.80 & 0.94 & \textbf{0.86} & \textbf{0.96}\\
speaker    & 0.73 & 0.75 & 0.89 & \textbf{0.79} & \textbf{0.90}\\
table      & 0.55 & 0.56 & 0.85 & \textbf{0.69} & \textbf{0.90}\\
watercraft & 0.65 & 0.69 & 0.90 & \textbf{0.74} & \textbf{0.92}\\ 
\hline
mean       & 0.63 & 0.67 & 0.88 & \textbf{0.74} & \textbf{0.91}\\ 
\hline
\end{tabular}}
\caption{Reconstruction IoU (\footnotesize{$\uparrow$}) on ShapeNet from 3000 points voxelized
at resolution 128³ (right column), and 300 points voxelized at resolution 32³ (left column).}
\label{tab:iou}
\end{table}

\begin{table}[h!]
\centering
\scalebox{0.8}{\begin{tabular}{|c|C{45pt}|C{30pt}C{30pt}|C{30pt}C{30pt}|}
\cline{2-6}
\multicolumn{1}{c|}{} & MetaSDF \footnotesize{(3k pts)} & \multicolumn{2}{c|}{\multirow{2}{*}{IF-Nets}} &  \multicolumn{2}{c|}{\multirow{2}{*}{\textbf{Ours}}}\\
\hline
airplane   & 0.360 & 0.097 & 0.013  & \textbf{0.067} & \textbf{0.006}\\
bench      & 0.407 & 0.369 & 0.015  & \textbf{0.262} & \textbf{0.010}\\
cabinet    & 0.463 & 0.401 & 0.123  & \textbf{0.234} & \textbf{0.114}\\
car        & 0.207 & 0.139 & 0.020  & \textbf{0.110} & \textbf{0.019}\\
chair      & 0.657 & \textbf{0.215} & 0.021  & 0.289 & \textbf{0.016}\\
display    & 0.682 & 0.151 & 0.019  & \textbf{0.086} & \textbf{0.014}\\
lamp       & 2.009 & \textbf{0.484} & 0.027  & 0.655 & \textbf{0.022}\\
phone      & 0.156 & 0.086 & 0.010  & \textbf{0.038} & \textbf{0.008}\\
rifle      & 0.212 & 0.066 & 0.013  & \textbf{0.044} & \textbf{0.006}\\
sofa       & 0.227 & 0.186 & 0.016  & \textbf{0.101} & \textbf{0.014}\\
speaker    & 0.645 & 0.339 & 0.101  & \textbf{0.295} & \textbf{0.084}\\
table      & 0.550 & 0.169 & 0.021 & \textbf{0.111} & \textbf{0.015}\\
watercraft & 0.394 & \textbf{0.308} & 0.013 & 0.422 & \textbf{0.010}\\ 
\hline
mean       & 0.458 & 0.232 & 0.032 & \textbf{0.209} & \textbf{0.026}\\
\hline
\end{tabular}}
\caption{Reconstruction $\text{CD}_\text{2} \times 10^{-3}$ (\footnotesize{$\downarrow$}) on ShapeNet from 3000 points voxelized
at resolution 128³ (right column), and 300 points voxelized at resolution 32³ (left column).}
\label{tab:cd2}
\end{table}

\begin{table}[h!]
\centering
\scalebox{0.8}{\begin{tabular}{|c|C{45pt}|C{30pt}C{30pt}|C{30pt}C{30pt}|}
\cline{2-6}
\multicolumn{1}{c|}{} & MetaSDF \footnotesize{(3k pts)} & \multicolumn{2}{c|}{\multirow{2}{*}{IF-Nets}} &  \multicolumn{2}{c|}{\multirow{2}{*}{\textbf{Ours}}}\\
\hline
airplane   & 0.099 & 0.061 & 0.021 & \textbf{0.047} & \textbf{0.019}\\
bench      & 0.120 & 0.097 & 0.028 & \textbf{0.072} & \textbf{0.024}\\
cabinet    & 0.122 & 0.116 & 0.053 & \textbf{0.083} & \textbf{0.048}\\
car        & 0.095 & 0.082 & 0.033 & \textbf{0.067} & \textbf{0.032}\\
chair      & 0.156 & 0.096 & 0.033 & \textbf{0.076} & \textbf{0.030}\\
display    & 0.133 & 0.091 & 0.032 & \textbf{0.061} & \textbf{0.027}\\
lamp       & 0.248 & \textbf{0.120} & 0.030 & 0.121 & \textbf{0.027}\\
phone      & 0.067 & 0.073 & 0.027 & \textbf{0.040} & \textbf{0.021}\\
rifle      & 0.081 & 0.059 & 0.020 & \textbf{0.043} & \textbf{0.016}\\
sofa       & 0.100 & 0.090 & 0.032 & \textbf{0.061} & \textbf{0.028}\\
speaker    & 0.151 & 0.116 & 0.047 & \textbf{0.093} & \textbf{0.045}\\
table      & 0.089 & 0.127 & 0.034 & \textbf{0.063} & \textbf{0.029}\\
watercraft & 0.119 & 0.094 & 0.026 & \textbf{0.080} & \textbf{0.022}\\ 
\hline
mean       & 0.123 & 0.091 & 0.032 & \textbf{0.070} & \textbf{0.028}\\
\hline
\end{tabular}}
\caption{Reconstruction $\text{CD}_\text{1} \times 10^{-1}$ (\footnotesize{$\downarrow$}) on ShapeNet from 3000 points voxelized
at resolution 128³ (right column), and 300 points voxelized at resolution 32³ (left column).}
\label{tab:cd1}
\end{table}

\end{document}